\documentclass[11pt,a4paper]{googlecloud}

\usepackage[authoryear,sort&compress,round]{natbib}
\usepackage{booktabs}
\usepackage{tabularx}
\usepackage{array}
\usepackage{multirow}
\usepackage{makecell}
\usepackage{graphicx}
\usepackage{xcolor}
\usepackage{colortbl}
\usepackage{enumitem}
\usepackage{algorithm}
\usepackage{algpseudocode}
\usepackage{placeins}
\usepackage{amsmath}
\usepackage{amssymb}
\usepackage{microtype}
\usepackage{url}
\usepackage{xspace}
\usepackage{comment}
\usepackage{fvextra}

\hypersetup{
  colorlinks=true,
  citecolor=blue,
  linkcolor=red,
  urlcolor=blue
}

\newcommand{\method}{\textsc{HDSO}\xspace}
\newcommand{\skillos}{\textsc{SkillOS}\xspace}
\newcommand{\transfer}{frozen transfer\xspace}
\newcommand{\todo}[1]{\textcolor{red!70!black}{[TBD: #1]}}

\newcolumntype{Y}{>{\raggedright\arraybackslash}X}
\newcolumntype{Z}{>{\centering\arraybackslash}X}

\title{Hypothesis-Driven Skill Optimization for LLM Agents}
\renewcommand{\today}{}
\author[1]{Shang Fangxin}
\author[1]{Yehui Yang}
\affil[1]{AI Lab, Qifu Technology, Beijing, China}
\correspondingauthor={Yehui Yang, Contact: \href{mailto:yangyehuisw@126.com}{yangyehuisw@126.com}}

\begin{abstract}
External skills can improve action-oriented LLM agents without changing model weights, but persistent skill updates are risky when they are distilled from sparse or noisy trajectories. A plausible reflection may encode a useful procedure, a spurious shortcut, or a rule that the target executor cannot reliably follow. We propose Hypothesis-Driven Skill Optimization (\method), a train-free framework in which both the skill curator and the agent executor are frozen inference endpoints. The curator observes executor traces, proposes a falsifiable hypothesis with an explicit validation plan, instantiates the hypothesis as a candidate skill package, validates the package through paired control/treatment executions, reviews behavior differences, and consolidates only supported candidates into an approved repository. The executor consumes approved skills through progressive disclosure, preserving the executor-only path when no skill is selected. On ALFWorld, \method improves executor-only baselines by +6.9 Avg.~SR points for Qwen3-8B and +4.0 points for Qwen3.6-27B. Under 20\% randomly flipped success/failure feedback during skill discovery and validation, \method preserves a +7.1-point gain for Qwen3-8B. Transfer and heterogeneous-pair diagnostics further show that validated repositories can be useful beyond the run that produced them, but cross-model curation succeeds only when curator diagnosis, executor capability, and validation evidence align. \method provides an auditable skill lifecycle for frozen action agents rather than an unconstrained memory accumulation procedure.
\end{abstract}

\begin{document}
\maketitle

\section{Introduction}

LLM agents are increasingly deployed as long-running systems that interact with environments, call tools, and face recurring task patterns. In action-oriented settings, many failures are procedural: the executor repeats an unproductive search, chooses an invalid action format, uses an object before satisfying a precondition, or fails to apply a known state transformation. A natural way to improve such agents without changing model weights is to maintain external skills: reusable procedures, invocation rules, output repairs, or small deterministic helpers that can be inspected and revised. This is attractive in private or enterprise deployments where the serving model is fixed because of cost, compliance, latency, or vendor constraints.

The central difficulty is not writing a skill, but deciding when a written skill should become durable agent knowledge. A failed trajectory may reflect a missing domain rule, weak exploration, an invalid action format, a poor invocation condition, an executor limitation, or unreliable success/failure feedback. If every reflection is appended to the prompt, the repository can accumulate over-general and contradictory rules. If skill optimization requires training a curator model, the approach becomes harder to deploy in inference-only settings.

We argue that persistent skill updates should be treated as hypotheses. A candidate skill should state what behavior it is expected to change, when it applies, what evidence motivated it, what risks it introduces, and what observations would falsify it. The system should then test the candidate prospectively: compare the current repository against the repository plus the candidate on matched tasks, inspect behavior differences, and promote the candidate only when the evidence supports the proposed mechanism.

\begin{figure}[t]
  \centering
  \includegraphics[width=\linewidth]{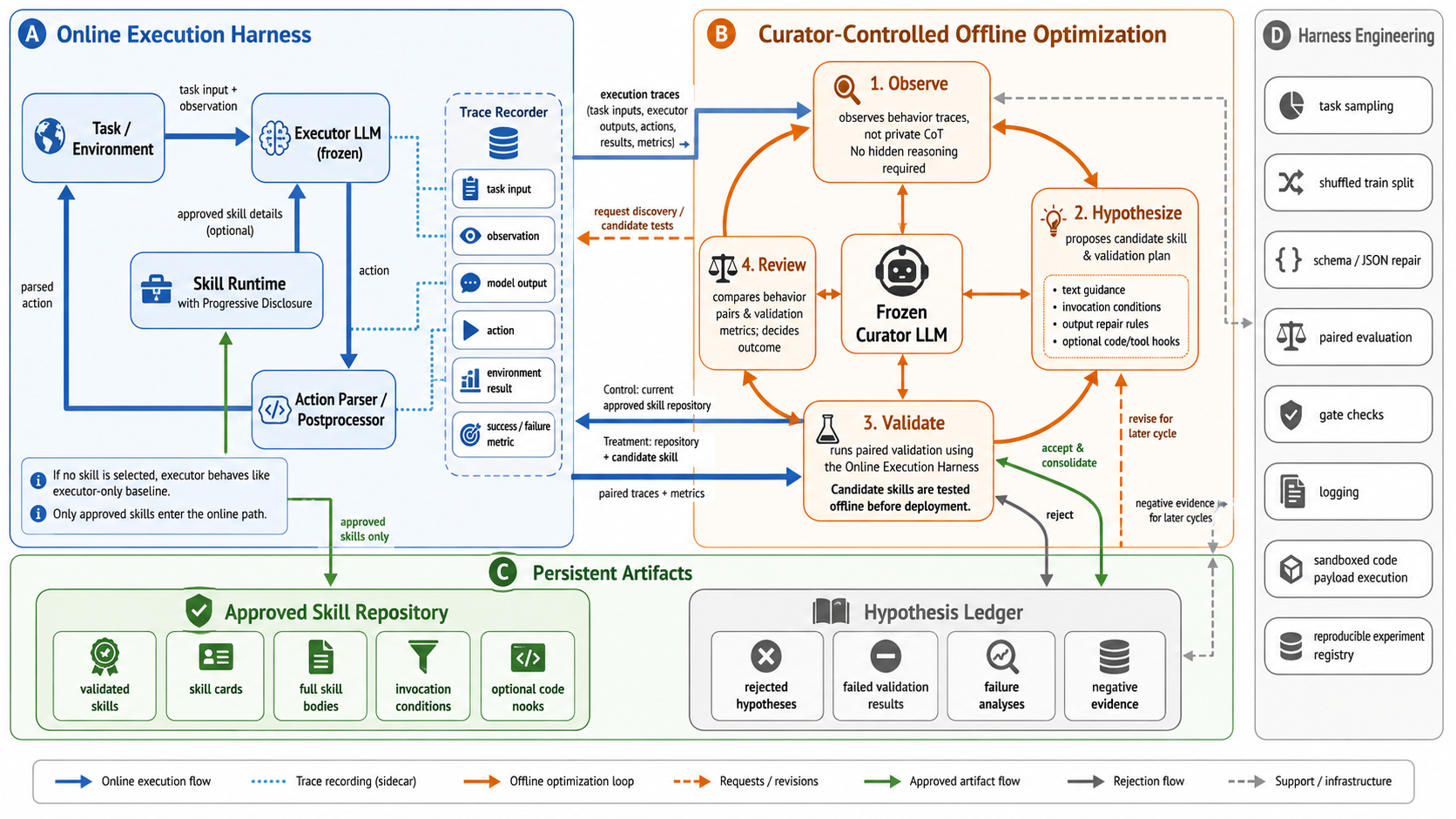}
  \caption{\method overview. Online execution uses a frozen executor and records task inputs, observations, actions, and outcomes. Offline optimization is controlled by a frozen curator that observes traces, proposes falsifiable skill hypotheses, validates them through paired control/treatment executions, and reviews the resulting behavior differences. Only approved skills enter the runtime repository; rejected hypotheses remain as auditable negative evidence.}
  \label{fig:overview}
\end{figure}

We instantiate this idea as Hypothesis-Driven Skill Optimization (\method), summarized in Figure~\ref{fig:overview}. \method decouples a frozen \emph{skill curator} from a frozen \emph{agent executor}. The curator observes compact executor traces and outcome feedback, proposes structured hypotheses, validates candidate skills by paired execution, reviews behavior differences, and maintains a hypothesis ledger. The executor consumes approved skills through progressive disclosure: it first sees compact skill cards and requests full details only when a skill appears relevant. When no skill is available or selected, the executor uses the same action loop as the executor-only baseline.

The train-free design targets a practical operating mode. A deployed executor can serve tasks online, while logs are used asynchronously to improve an external skill repository during low-load periods. The update does not require fine-tuning, reward-model training, or modifying the serving endpoint. It produces artifacts that can be audited: approved skills, rejected hypotheses, validation pairs, and failure attributions.

The paper makes three contributions.
\begin{enumerate}[leftmargin=*]
  \item We formulate persistent skill updates as a hypothesis lifecycle that separates observation, proposal, validation, review, consolidation, and rejection.
  \item We instantiate this lifecycle in a train-free curator-executor framework with paired validation and progressive skill disclosure; both curator and executor parameters remain fixed.
  \item We evaluate \method on ALFWorld, showing positive gains over executor-only baselines, robustness under noisy success/failure feedback during discovery and validation, transfer evidence for validated repositories, and diagnostics for when cross-model curation fails.
\end{enumerate}

\section{Related Work}

\paragraph{Self-evolving agents and external skills.}
External skills are a common substrate for agent self-improvement. Voyager-style agents store executable skills for future environment interaction \citep{voyager2023}. Tool-creation work studies how LLMs synthesize reusable functions or APIs for subsequent tasks \citep{creator2023,toolmakers2023}. Recent self-evolving systems optimize skill content, skill invocation, or skill-aware reflection \citep{skillopt2026,skillevolver2026,embodiskill2026,sega2026}. \method shares the premise that persistent artifacts can improve frozen agents, but focuses on the admission rule: a skill is not accepted because it sounds plausible, but because a falsifiable hypothesis survives prospective validation.

\paragraph{Automatic skill optimization.}
\skillos is the closest comparison because it separates a skill curator from an agent executor and studies skill repositories as the substrate for self-evolving agents \citep{skillos2026}. Its full system includes a learned curator, but its reported \textsc{SkillOS}-base setting is closer to the comparison we use: skills are produced without the trained curator policy. SkillOpt also treats executive strategy as an optimizable skill object for self-evolving agents \citep{skillopt2026}. We therefore do not frame \method as merely ``untrained SkillOS.'' The difference is the update rule. \method requires each proposed skill to pass an explicit observe--hypothesize--validate--review--consolidate lifecycle, producing validation artifacts and rejected-hypothesis records rather than only an accumulated skill set.

\paragraph{Reflection and memory.}
Reflection and experiential learning methods convert completed trajectories into verbal lessons or memories \citep{reflexion2023,expel2023}. Context-evolution methods update an operating context directly from new evidence \citep{ace2026}. ReasoningBank is a strong memory baseline that reports large gains from distilled reasoning memories \citep{reasoningbank2026}. Our claim is not that hypothesis-driven skills dominate all memory methods on raw score. Instead, \method targets trust management for persistent agent knowledge: each approved skill has scope, expected effect, validation evidence, and rollback rationale.

\section{Method}

\subsection{Problem Setup}

Let $E_\theta$ be a frozen executor LLM and $C_\phi$ a frozen curator LLM, with fixed parameters $\theta$ and $\phi$. Let $\mathcal{D}_{train}$ be a task stream used for skill optimization, $\mathcal{D}_{eval}$ a held-out evaluation stream, and $\mathcal{S}$ the approved skill repository. For a task $x$, the executor policy conditioned on repository $\mathcal{S}$ samples a trajectory
\begin{equation}
\tau=(o_1,a_1,\ldots,o_T,a_T), \qquad \tau \sim \pi_{E_\theta}(\cdot \mid x,\mathcal{S}),
\end{equation}
where $o_t$ is the environment observation at step $t$ and $a_t$ is the executor action. The trajectory induces metrics $m(\tau)=(y(\tau), v(\tau), \ell(\tau))$, where $y\in\{0,1\}$ is task success, $v$ counts invalid actions, and $\ell$ is the number of environment steps. The curator observes task inputs, executor actions, environment observations, outcomes, approved skills, and validation summaries, but not executor private reasoning.

The goal is to update only $\mathcal{S}$ so that expected evaluation utility improves:
\begin{equation}
\max_{\mathcal{S}'}\; \mathbb{E}_{x\sim\mathcal{D}_{eval}}\,
\mathbb{E}_{\tau\sim \pi_{E_\theta}(\cdot\mid x,\mathcal{S}')}\left[U(\tau)\right]
\quad\text{s.t.}\quad \theta,\phi\ \text{remain fixed},
\end{equation}
where $U$ is primarily success rate with invalid-action and step-count diagnostics. The executor-only baseline is the special case $\mathcal{S}=\emptyset$.

\method is built around four invariants. First, the no-skill executor path is the executor-only baseline. Second, candidate skills are structured artifacts with scope and falsification conditions. Third, validation is prospective and paired: control and treatment arms run on the same task indices. Fourth, rejected candidates remain in a hypothesis ledger, so future cycles can avoid rediscovering the same failed rule or can narrow a promising direction.

\subsection{Executor with Skill Conditioning}

The executor is a standard ReAct-style environment policy augmented with an optional skill interface. At each step, the runtime retrieves at most a small number of compact skill cards from $\mathcal{S}$ using the task, current observation, and visible interaction history. The executor may either emit an environment action directly or request the full content of one relevant skill. If it requests a skill, the runtime discloses only that skill and then asks for the next action. This interface is a delivery mechanism rather than a new learning algorithm; its role is to make approved skills available without appending the entire repository to every prompt.

This design preserves the executor-only baseline. When $\mathcal{S}=\emptyset$, or when the executor does not request a skill, the action-generation path is the same as the no-skill ReAct executor. Skill usage is also observable: each request for a full skill body becomes an uptake event recorded in the trajectory. Optional skill payloads may provide restricted pre-prompt or post-action helpers, but they are skill-owned helpers rather than a general tool-call API, and they must preserve baseline behavior when uncertain.

\begin{algorithm}[H]
\caption{Skill-Conditioned ReAct Executor}
\label{alg:executor}
\begin{algorithmic}[1]
\Require Task $x$, instruction $I$, environment, repository $\mathcal{S}$, selector $R$, executor $E_\theta$
\Ensure Trajectory $\tau$ and metrics $m$
\State Initialize history $h \gets \emptyset$
\For{$t=1,\ldots,T$}
  \State Observe $o_t$ and retrieve compact cards $K_t \gets R(\mathcal{S}, x, o_t, h)$
  \State Query $E_\theta$ with $(I,o_t,h,K_t)$
  \If{the output requests skill $s\in K_t$}
    \State Disclose $s$ and query $E_\theta$ again for one action
  \EndIf
  \State Parse one action $a_t$ and apply any safe skill-owned payload transform
  \State Execute the action; append transition to $h$
\EndFor
\State Compute metrics $m=(y,v,\ell)$ from the completed trajectory
\State \Return $\tau,m$
\end{algorithmic}
\end{algorithm}

\subsection{Curator Optimization Loop}

\begin{figure}[t]
  \centering
  \includegraphics[width=0.92\linewidth]{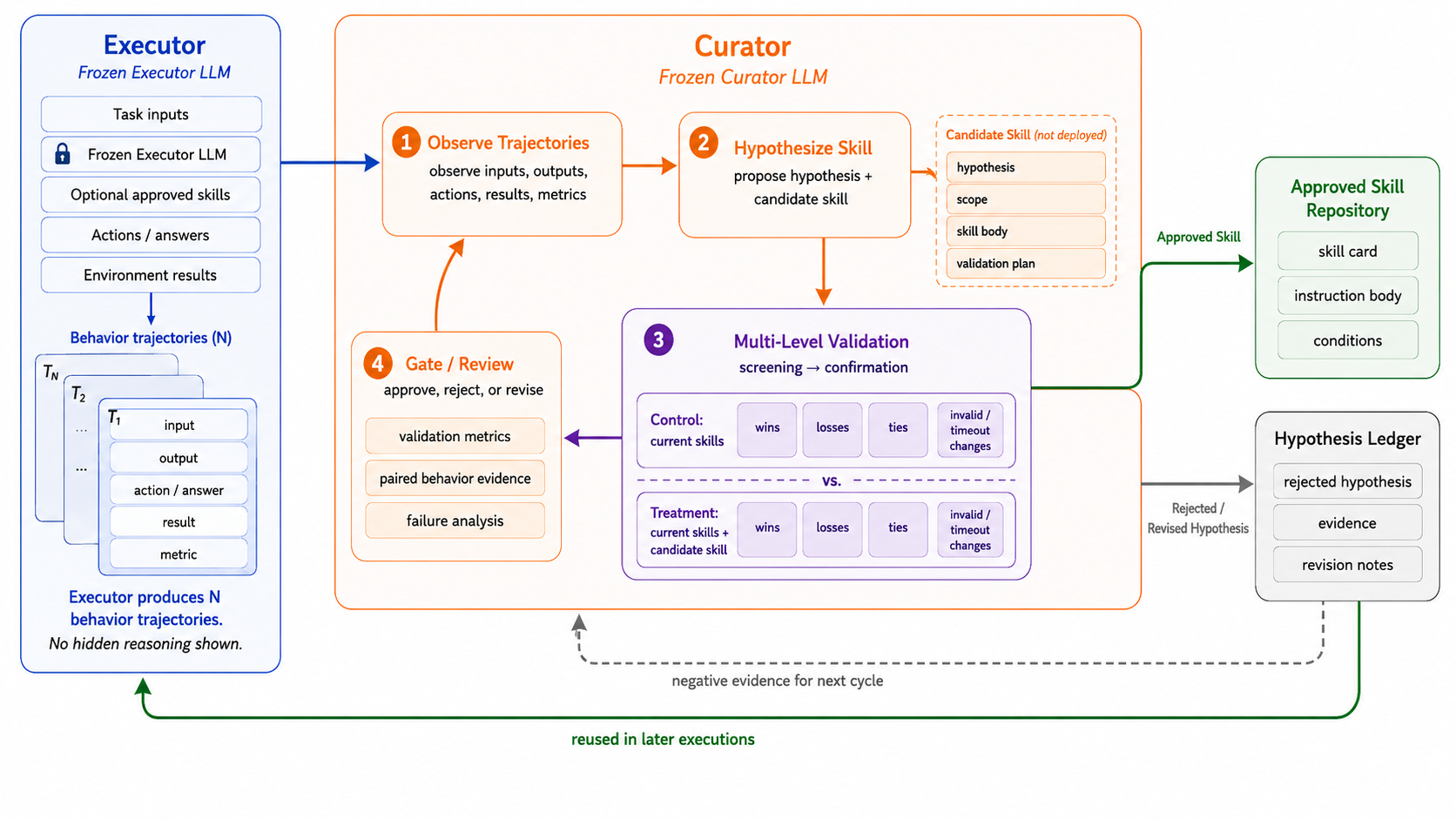}
  \caption{Detailed \method lifecycle. The curator turns discovery traces into candidate skill hypotheses, evaluates each candidate through staged paired validation, reviews treatment-specific behavior changes, and consolidates only candidates with supported mechanisms and acceptable guardrail risk.}
  \label{fig:method}
\end{figure}

\method maintains three state objects: an approved repository $\mathcal{S}$, a hypothesis bank $\mathcal{H}$, and an evidence ledger $\mathcal{L}$. Figure~\ref{fig:method} shows the optimization loop. Each cycle begins with discovery traces generated by the current executor and repository. The curator receives compact behavior evidence rather than executor private reasoning: task instructions, observations, admissible-action samples, chosen actions, final states, invalid-action markers, success/failure labels, repeated actions, task-type statistics, approved skill summaries, rejected hypothesis summaries, and later paired validation behavior.

A curator proposal is a structured hypothesis with a candidate skill. It states the behavioral claim, supporting evidence, intended scope, expected metric or behavior change, risks and rollback conditions, and the skill artifact to test. The skill artifact contains the text body, invocation conditions, retrieval metadata, and optional payloads; the implementation schema is given in Appendix~\ref{app:prompt}. Candidate validation is staged for efficiency. Screening-small checks for an initial positive effect, screening-medium removes fragile candidates, and confirmation uses independent task indices.

For a validation set $V$, the control arm uses $\mathcal{S}$ and the treatment arm uses $\mathcal{S}\cup\{s\}$ on the same tasks:
\begin{equation}
\tau_i^{0}\sim \pi_{E_\theta}(x_i,\mathcal{S}),\qquad
\tau_i^{1}\sim \pi_{E_\theta}(x_i,\mathcal{S}\cup\{s\}).
\end{equation}
Let $b=\sum_i \mathbb{1}[y(\tau_i^{1})=1,y(\tau_i^{0})=0]$ be treatment-only wins and $c=\sum_i \mathbb{1}[y(\tau_i^{1})=0,y(\tau_i^{0})=1]$ be control-only wins. The core effect statistic is
\begin{equation}
\mathrm{net}(s;V)=b-c,\qquad
\Delta_{\mathrm{SR}}(s;V)=\frac{1}{|V|}\sum_i\left(y(\tau_i^{1})-y(\tau_i^{0})\right).
\end{equation}

The promotion gate records these paired outcomes, success delta, step delta, invalid-action delta, guardrail regressions, per-scope regressions, skill uptake, and an exact one-sided McNemar/binomial statistic. We use the p-value
\begin{equation}
p_{\mathrm{binom}}=\Pr\!\left[\mathrm{Binom}(b+c,0.5)\ge b\right]
\end{equation}
as a diagnostic of paired-evidence strength. It is logged for analysis and audit, but it is not a hard promotion threshold in the main experiments. Promotion depends on minimum effect, guardrails, invalid-action behavior, per-scope regression checks, and curator review. The review stage classifies each candidate as a validated direction, a promising direction that needs revision, a wrong direction, or insufficient evidence. Accepted candidates are consolidated into $\mathcal{S}$; rejected candidates remain in $\mathcal{H}$ with failure attribution.

\begin{algorithm}[H]
\caption{Hypothesis-Driven Skill Curator}
\label{alg:curator}
\begin{algorithmic}[1]
\Require Executor $E_\theta$, curator $C_\phi$, train pool $\mathcal{D}_{train}$, repository $\mathcal{S}$, hypothesis bank $\mathcal{H}$, cycle budget $N_{\mathrm{cyc}}$
\Ensure Updated repository $\mathcal{S}$, hypothesis bank $\mathcal{H}$, evidence ledger $\mathcal{L}$
\For{$r=1,\ldots,N_{\mathrm{cyc}}$}
  \State Collect discovery traces $D_r$ with $E_\theta$ and $\mathcal{S}$
  \State Ask $C_\phi$ for candidate hypotheses $\{h_j\}$ and their skills $\{s_j\}$
  \For{each candidate $h_j$ with skill $s_j$}
    \State Run paired screening stages for $\mathcal{S}$ vs. $\mathcal{S}\cup\{s_j\}$
    \If{any screening gate fails}
      \State Store screening evidence in $\mathcal{H}$ and $\mathcal{L}$; \textbf{continue}
    \EndIf
    \State Run paired confirmation on independent train tasks
    \State Ask $C_\phi$ to review behavior pairs and assign a direction label
    \If{metric gate and review criteria pass}
      \State $\mathcal{S}\gets \mathcal{S}\cup\{s_j\}$
    \Else
      \State Store rejection or revision evidence in $\mathcal{H}$ and $\mathcal{L}$
    \EndIf
  \EndFor
\EndFor
\State \Return $\mathcal{S},\mathcal{H},\mathcal{L}$
\end{algorithmic}
\end{algorithm}

\FloatBarrier

\section{Experiments}

The experiments separate three questions. First, can a frozen curator improve an executor powered by the same LLM? Second, once a repository is validated, can the skills transfer to another executor without an active curator? Third, can one frozen LLM autonomously curate skills for an executor powered by a different LLM? The first question tests homogeneous curator-executor optimization; the latter two diagnose transfer and heterogeneous-pair generalization.

\subsection{Setup}

We evaluate on ALFWorld \citep{alfworld2020}. HDSO optimization uses tasks from ALFWorld's official train split: the curator observes train-split trajectories, proposes skills, and validates candidates on held-out pools sampled from the same split. We report metrics on the ALFWorld evaluation set following the standard evaluation protocol used by SkillOS and related ALFWorld agent studies. We report Avg.~SR and Avg.~Steps, with Avg.~Steps measured as environment steps per episode. To reduce sensitivity to task ordering and stochastic generation, each run fixes the seed for train-task shuffling, validation-pool construction, and model sampling; unless stated otherwise, results are averaged over three seeds.

The curator receives compact evidence extracted from executor runs. A discovery trace contains the task instruction, environment observations, executed actions, final success/failure outcome, and compact failure evidence. ALFWorld HDSO uses $N_{\mathrm{cyc}}=3$ curator cycles. At the beginning of each cycle, discovery collects three fresh traces per task type; later cycles use \emph{discovery refill} to add non-overlapping traces before proposal. Each proposal call can return up to eight candidate hypotheses, each paired with a candidate skill, an intended scope, expected effect, validation plan, and guardrails.

Candidate skills are validated prospectively through paired control/treatment runs on train-split tasks. The control run uses the current approved repository, while the treatment run adds the candidate under review. Validation proceeds through screening-small, screening-medium, and confirmation. Screening-small uses four target tasks and two guardrail tasks; screening-medium uses eight target tasks and four guardrail tasks; confirmation uses 32 target tasks and 12 guardrail tasks. We define net wins as treatment-only successes minus control-only successes, and guardrail regressions as paired losses outside the candidate's intended scope. Screening requires at least one net win and at least a 0.05 success-rate improvement, with zero guardrail regression in screening-small and at most one afterward; confirmation uses the same success-rate threshold and at most one guardrail regression. A one-sided McNemar/binomial p-value at $\alpha=0.05$ is logged as an audit statistic rather than a hard promotion gate.

Executor episodes are capped at 30 environment steps. The skill selector exposes at most three compact skill cards per step, and full skill bodies are disclosed only when requested by the executor. Curator calls use temperature 0.3, top-$p=1.0$, native model reasoning, and 12,288 output tokens. Executor sampling follows the corresponding model configuration; all HDSO and executor-only runs share the same ReAct environment interface. Models include Qwen-family checkpoints \citep{qwen3technical2025}: Qwen3-8B, Qwen3.5-9B, and Qwen3.6-27B. For active HDSO runs, $C\rightarrow E$ denotes curator model $C$ and executor model $E$; in \transfer experiments, a Repository Source of $C\rightarrow E$ denotes a repository produced by that active run and reused for a later executor. Qwen3.5-9B is retained for executor baselines and heterogeneous-pair diagnostics. External SkillOS and ReasoningBank values are cited from prior work, and cross-paper comparisons emphasize deltas relative to each method family's executor-only baseline.

\subsection{ALFWorld Main Results}

\begin{table}[t]
\centering
\scriptsize
\begin{tabularx}{\linewidth}{ZZZZZZ}
\toprule
Method & Curator & Executor & Avg.~SR & Delta & Avg.~Steps \\
\midrule
Executor-only & none & Qwen3-8B & 40.5 & 0.0 & 22.3 \\
Executor-only & none & Qwen3.5-9B & 44.3 & 0.0 & 21.9 \\
Executor-only & none & Qwen3.6-27B & 61.7 & 0.0 & 17.9 \\
\midrule
MemP & Qwen3-8B & Qwen3-8B & -- & +1.8 & 21.0 \\
\textsc{SkillOS}-base & Qwen3-8B & Qwen3-8B & -- & +5.2 & 20.4 \\
ReasoningBank & Qwen3-8B & Qwen3-8B & -- & +7.8 & 20.1 \\
\midrule
\method & Qwen3-8B & Qwen3-8B & 47.4 & +6.9 & 20.9 \\
\method & Qwen3-8B & Qwen3.5-9B & 52.1 & +7.8 & 20.8 \\
\method & Qwen3.6-27B & Qwen3.6-27B & 65.7 & +4.0 & 17.3 \\
\bottomrule
\end{tabularx}
\caption{Main ALFWorld results. Avg.~SR is computed over 140 episodes per seed; Avg.~Steps is normalized per episode. External baselines are interpreted primarily by their reported gains over their own executor-only baselines because serving stacks and prompts are not byte-identical.}
\label{tab:main}
\end{table}

Table~\ref{tab:main} evaluates whether \method can improve frozen ALFWorld agents under both homogeneous and heterogeneous curator-executor pairings. In the homogeneous setting, the curator and executor are powered by the same LLM, so the result isolates train-free self-improvement with fixed curator and executor parameters. Qwen3-8B improves from 40.5 to 47.4 Avg.~SR (+6.9), and Qwen3.6-27B improves from 61.7 to 65.7 (+4.0). Avg.~Steps also decreases in both rows, indicating that HDSO improves success while keeping the interaction budget efficient.

The heterogeneous 8B$\rightarrow$9B result shows that the active HDSO optimization loop can operate when the curator and executor use different LLMs. A Qwen3-8B curator raises the Qwen3.5-9B executor from 44.3 to 52.1 Avg.~SR, the largest HDSO gain in Table~\ref{tab:main}. This supports the view that the curator can diagnose another executor's failures and produce a target-executable intervention, not only optimize the behavior of an executor powered by the same LLM. At the same time, heterogeneous curation should be treated as a pairwise alignment problem rather than a monotonic function of curator scale. We return to difficult heterogeneous pairings in the discussion.

The external rows position \method against prior skill and memory approaches. Because serving stacks, prompts, and baselines are not byte-identical across papers, the most stable comparison is the reported delta over each method's executor-only baseline. HDSO's homogeneous 8B gain (+6.9) exceeds the reported \textsc{SkillOS}-base gain (+5.2) and MemP gain (+1.8), while the heterogeneous 8B$\rightarrow$9B gain (+7.8) matches the reported ReasoningBank delta. The comparison positions HDSO as a train-free, validation-gated skill lifecycle that reaches competitive improvements while retaining explicit evidence, rejection records, and rollback context for each admitted skill.

\begin{table}[t]
\centering
\scriptsize
\begin{tabularx}{\linewidth}{
>{\hsize=1.80\hsize\raggedright\arraybackslash}X
>{\hsize=0.50\hsize\centering\arraybackslash}X
>{\hsize=0.50\hsize\centering\arraybackslash}X
>{\hsize=0.55\hsize\centering\arraybackslash}X
>{\hsize=0.75\hsize\centering\arraybackslash}X
>{\hsize=0.90\hsize\centering\arraybackslash}X
>{\hsize=2.00\hsize\raggedright\arraybackslash}X}
\toprule
Task type & Baseline & \method & Delta & HDSO-wins & Baseline-wins & Dominant mechanism \\
\midrule
\texttt{look\_\allowbreak at\_\allowbreak obj} & 46.2 & 53.8 & +7.6 & 1 & 0 & acquisition before illumination \\
\texttt{pick\_\allowbreak and\_\allowbreak place} & 85.7 & 88.6 & +2.9 & 2 & 1 & search and placement checks \\
\texttt{pick\_\allowbreak clean\_\allowbreak then\_\allowbreak place} & 44.4 & 44.4 & 0.0 & 2 & 2 & progress tracking \\
\texttt{pick\_\allowbreak cool\_\allowbreak then\_\allowbreak place} & 48.0 & 56.0 & +8.0 & 2 & 0 & progress tracking \\
\texttt{pick\_\allowbreak heat\_\allowbreak then\_\allowbreak place} & 43.8 & 62.5 & +18.7 & 4 & 1 & direct heating action selection \\
\texttt{pick\_\allowbreak two\_\allowbreak obj} & 75.0 & 75.0 & 0.0 & 1 & 1 & multi-object tracking \\
\bottomrule
\end{tabularx}
\caption{Task-type behavior analysis for the homogeneous Qwen3.6-27B curator-executor setting. HDSO-wins counts paired episodes solved only by \method, while Baseline-wins counts paired episodes solved only by the executor-only baseline. The largest gain comes from \texttt{pick\_heat\_then\_place}, where approved skills change a concrete action-selection mechanism.}
\label{tab:tasktype}
\end{table}

Table~\ref{tab:tasktype} provides mechanism-level evidence for the main result. Gains concentrate in task families with recurring procedural failures, especially heating, cooling, and look-at tasks. The largest improvement appears in \texttt{pick\_heat\_then\_place}, where the approved skill changes a concrete action-selection rule around the heating transformation. This pattern is consistent with the intended role of a skill: reusable procedural knowledge that applies across episodes, rather than a one-off reflection. The unchanged clean and multi-object rows also explain why HDSO requires scoped validation and guardrails; persistent skills must be admitted with evidence about where they help and where invocation remains limited.

\subsection{Skill Delivery}

\begin{table}[t]
\centering
\scriptsize
\begin{tabularx}{\linewidth}{
>{\hsize=1.10\hsize\centering\arraybackslash}X
>{\hsize=1.60\hsize\centering\arraybackslash}X
>{\hsize=1.00\hsize\centering\arraybackslash}X
>{\hsize=1.65\hsize\raggedright\arraybackslash}X
>{\hsize=0.50\hsize\centering\arraybackslash}X
>{\hsize=0.45\hsize\centering\arraybackslash}X
>{\hsize=0.75\hsize\centering\arraybackslash}X}
\toprule
Method & Repository Source & Executor & Delivery Interface & Avg.~SR & Delta & Avg.~Steps \\
\midrule
\transfer & \makecell[c]{HDSO(8B$\rightarrow$8B)} & Qwen3-8B & Full skill context & 42.1 & +1.6 & 21.8 \\
\transfer & \makecell[c]{HDSO(27B$\rightarrow$27B)} & Qwen3-8B & Progressive disclosure & 46.2 & +5.7 & 21.1 \\
\transfer & \makecell[c]{HDSO(27B$\rightarrow$27B)} & Qwen3.6-27B & Full skill context & 62.9 & +1.2 & 17.3 \\
\bottomrule
\end{tabularx}
\caption{\transfer evaluation of approved HDSO skill repositories with the curator disabled at evaluation time. The table compares two delivery interfaces and tests whether a repository discovered in one curator-executor configuration remains useful for another executor. Repository Source identifies the active HDSO run that produced the approved repository. Delta is computed against the corresponding executor-only baseline. Full skill context places the complete approved repository in the executor context throughout evaluation, whereas progressive disclosure exposes compact skill cards and reveals full skill details only on request.}
\label{tab:transfer}
\end{table}

Table~\ref{tab:transfer} examines how an approved repository should be delivered
to an executor after the active HDSO loop has produced it. In \transfer, the
curator is disabled at evaluation time: the repository is fixed, no new
discovery or validation occurs, and the executor only consumes the approved
skills. We compare two delivery interfaces. Full skill context places the
complete repository in the executor context throughout evaluation. Progressive
disclosure exposes compact skill cards first and reveals full skill content only
when the executor requests a specific skill.

Full skill context is a useful but limited baseline. With the repository
discovered in the homogeneous Qwen3-8B run, the Qwen3-8B executor improves from
40.5 to 42.1 Avg.~SR. With the repository discovered in the homogeneous
Qwen3.6-27B run, the Qwen3.6-27B executor improves from 61.7 to 62.9 Avg.~SR.
These gains show that the approved skill text carries actionable information
even when it is delivered as static context. However, the effect is modest,
suggesting that simply appending the full repository does not reliably make the
executor use the right skill at the right state.

Progressive disclosure provides a stronger use pattern in the available transfer
setting. Reusing the repository discovered in the homogeneous Qwen3.6-27B
curator-executor run with the Qwen3-8B executor improves Avg.~SR from 40.5 to
46.2, a gain of +5.7 points. This row supports two points. First, the repository
contains reusable procedural knowledge rather than only executor-specific prompt
artifacts. Second, skill delivery matters: exposing compact cards before full
skill bodies gives the executor a mechanism to select relevant skills without
unconditionally polluting every prompt with the entire repository.

Thus, Table~\ref{tab:transfer} complements the active HDSO results in
Table~\ref{tab:main}. Active HDSO is the optimization procedure that discovers
and validates skills; \transfer is a deployment mode for reusing the resulting
repository. The results indicate that approved repositories have standalone
value, but that value is realized more effectively when the executor accesses
skills through progressive disclosure rather than full skill context alone.

\subsection{Noisy Feedback}

We test whether \method can still produce useful skills when optimization feedback is unreliable. During discovery, screening, and confirmation, 20\% of success/failure labels are randomly flipped. Raw uncorrupted traces are stored separately, and final full evaluation uses true environment success. The curator prompt is informed that feedback may be noisy. This setting simulates corrupted optimization logs rather than a corrupted test set.

\begin{table}[t]
\centering
\scriptsize
\begin{tabularx}{\linewidth}{
>{\hsize=1.00\hsize\centering\arraybackslash}X
>{\hsize=1.00\hsize\centering\arraybackslash}X
>{\hsize=1.00\hsize\centering\arraybackslash}X
>{\hsize=1.65\hsize\raggedright\arraybackslash}X
>{\hsize=0.65\hsize\centering\arraybackslash}X
>{\hsize=0.50\hsize\centering\arraybackslash}X
>{\hsize=0.60\hsize\centering\arraybackslash}X}
\toprule
Method & Curator & Executor & Optimization Feedback & Avg.~SR & Delta & Avg.~Steps \\
\midrule
Executor-only & none & Qwen3-8B & none & 40.5 & 0.0 & 22.3 \\
\method & Qwen3-8B & Qwen3-8B & clean success/failure & 47.4 & +6.9 & 20.9 \\
\method & Qwen3-8B & Qwen3-8B & 20\% label flip & 47.6 & +7.1 & 20.7 \\
\bottomrule
\end{tabularx}
\caption{Feedback-corruption ablation on ALFWorld for the Qwen3-8B curator-executor setting. Label flipping is applied only to success/failure feedback observed during skill optimization; reported final evaluation uses uncorrupted environment success.}
\label{tab:noise}
\end{table}

Table~\ref{tab:noise} directly tests the deployment motivation. In real applications, success/failure may be the only available preference signal, but it may be produced by brittle scripts, delayed user behavior, or noisy downstream outcomes. Under 20\% label corruption during optimization, the Qwen3-8B setting reaches 47.6 Avg.~SR on clean final evaluation, compared with 47.4 under clean optimization feedback and 40.5 for the executor-only baseline. Avg.~Steps also remains comparable to the clean-feedback run. The result indicates that the HDSO lifecycle does not collapse when the optimization signal is moderately unreliable.

The mechanism is also important. The noisy-feedback run does not promote a broad set of generic reminders. It consolidates a narrow cleaning skill, \emph{Sinkbasin Navigation Optimization}, and the executor requests skills in only 28 of 420 evaluation episodes. Those requests concentrate on \texttt{pick\_clean\_then\_place}, the task family targeted by the approved skill, rather than spreading across unrelated task types. This behavior is consistent with the intended role of hypothesis-gated skill optimization: noisy observations may affect proposal search, but durable repository updates still pass through scoped validation, behavior review, and progressive disclosure.

We do not interpret label noise as beneficial. The 20\% label-flip result should be read as robustness evidence for this model pair and noise level, not as a general claim that corrupted feedback improves skill learning. The key point is that \method avoids the failure mode of directly absorbing every noisy reflection into the executor prompt; when evidence is unreliable, the lifecycle tends to admit narrow, testable skills rather than unconstrained memory.

\section{Discussion}

\method reframes skill optimization as hypothesis testing for frozen LLM agents. On ALFWorld, the homogeneous curator-executor results show that a frozen curator can improve an executor powered by the same LLM, while transfer experiments show that validated repositories can contain reusable external knowledge. Heterogeneous-pair diagnostics show the boundary: autonomous curation across model scales is harder than reusing a validated repository, because the curator must write skills that match the target executor's capabilities.

\textbf{Failure analysis.} Heterogeneous active curation is more demanding than \transfer because the curator must not only identify a reusable intervention, but also write one that the target executor can reliably execute. In the 27B$\rightarrow$8B run, logs show non-trivial skill uptake during screening, but small positive effects reversed or vanished in medium validation, indicating that plausible high-level procedures from a stronger curator may exceed the weaker executor's stable execution capability. The same issue appears when Qwen3.5-9B is used as the executor for skills proposed by Qwen3.6-27B: the accepted skill was locally valid for a narrow \texttt{look\_at\_obj} slice, but did not cover the executor's dominant failures on multi-step state-change tasks and therefore did not improve full evaluation. In the 8B$\rightarrow$27B run, the proposed rules were either already covered by the stronger executor's baseline behavior or too local to survive confirmation. Runs with Qwen3.5-9B as curator are also excluded from the main plan. Across 9B$\rightarrow$8B, 9B$\rightarrow$9B, and 9B$\rightarrow$27B attempts, verified serving and sampling settings still did not produce a stable approved repository, suggesting that this checkpoint is not reliable as an autonomous curator in our setting. These failures are useful because HDSO records where the lifecycle breaks, including proposal quality, executor uptake, action reliability, validation robustness, coverage mismatch, and review conservatism.

\textbf{Benchmark scope.} Our empirical evidence is limited to ALFWorld, a multi-step action and planning benchmark with explicit observations, environment actions, and terminal success feedback. This setting is well aligned with the \method abstraction: skills can modify state tracking, action selection, loop avoidance, and post-action checks, and their effects can be inspected through paired trajectories. The same lifecycle may extend naturally to other action-centric or planning-centric agent benchmarks where behavior changes are observable over multiple steps. It should not be assumed, however, that the same skill interface transfers unchanged to single-turn question answering or pure reasoning tasks. In those settings, the observable trajectory is often a one-shot answer trace rather than a sequence of environment decisions, so skill applicability, uptake, and validation evidence require a different adapter design.

\section{Conclusion}

\method casts persistent skill updates as a train-free hypothesis lifecycle for action-oriented LLM agents. A frozen curator observes executor traces, proposes falsifiable hypotheses and validation plans, instantiates candidate skills, and promotes only skills supported by paired execution evidence and behavior review. On ALFWorld, this lifecycle improves executor-only baselines, remains effective under noisy success/failure feedback during discovery and validation, and produces transfer and failure records that expose when cross-model curation does or does not align with the target executor. These results support \method as an auditable alternative to unvalidated prompt or memory accumulation for frozen action agents, while leaving broader question-answering and pure reasoning settings to future adapter designs.

\clearpage
\bibliographystyle{abbrvnat}
\bibliography{references}

\begin{thebibliography}{14}
\providecommand{\natexlab}[1]{#1}
\providecommand{\url}[1]{\texttt{#1}}
\expandafter\ifx\csname urlstyle\endcsname\relax
  \providecommand{\doi}[1]{doi: #1}\else
  \providecommand{\doi}{doi: \begingroup \urlstyle{rm}\Url}\fi

\bibitem[Cai et~al.(2023)]{toolmakers2023}
T.~Cai et~al.
\newblock Large language models as tool makers.
\newblock \emph{arXiv preprint arXiv:2305.17126}, 2023.

\bibitem[Jin et~al.(2026)Jin, Wang, and Zhang]{sega2026}
S.~Jin, L.~Wang, and Z.~Zhang.
\newblock Se-ga: Memory-augmented self-evolution for gui agents.
\newblock \emph{arXiv preprint arXiv:2605.16883}, 2026.

\bibitem[Ju et~al.(2026)]{embodiskill2026}
R.~Ju et~al.
\newblock Embodiskill: Skill-aware reflection for self-evolving embodied
  agents.
\newblock \emph{arXiv preprint arXiv:2605.10332}, 2026.

\bibitem[Ouyang et~al.(2026)Ouyang, Yan, Chen, Han, Wang, et~al.]{skillos2026}
S.~Ouyang, J.~Yan, Y.~Chen, R.~Han, Z.~Wang, et~al.
\newblock Skillos: Learning skill curation for self-evolving agents.
\newblock \emph{arXiv preprint arXiv:2605.06614}, 2026.

\bibitem[Ouyang et~al.(2025)]{reasoningbank2026}
S.~Ouyang et~al.
\newblock Reasoningbank: Scaling agent self-evolving with reasoning memory.
\newblock \emph{arXiv preprint arXiv:2509.25140}, 2025.

\bibitem[Qian et~al.(2023)]{creator2023}
C.~Qian et~al.
\newblock Creator: Tool creation for disentangling abstract and concrete
  reasoning of large language models.
\newblock \emph{arXiv preprint arXiv:2305.14318}, 2023.

\bibitem[Shinn et~al.(2023)]{reflexion2023}
N.~Shinn et~al.
\newblock Reflexion: Language agents with verbal reinforcement learning.
\newblock \emph{arXiv preprint arXiv:2303.11366}, 2023.

\bibitem[Shridhar et~al.(2021)Shridhar, Yuan, C{\^o}t{\'e}, Bisk, Trischler,
  and Hausknecht]{alfworld2020}
M.~Shridhar, X.~Yuan, M.-A. C{\^o}t{\'e}, Y.~Bisk, A.~Trischler, and
  M.~Hausknecht.
\newblock {ALFWorld}: Aligning text and embodied environments for interactive
  learning.
\newblock In \emph{International Conference on Learning Representations}, 2021.

\bibitem[Wang et~al.(2023)]{voyager2023}
G.~Wang et~al.
\newblock Voyager: An open-ended embodied agent with large language models.
\newblock \emph{arXiv preprint arXiv:2305.16291}, 2023.

\bibitem[Yang et~al.(2025)Yang, Li, Yang, Zhang, Hui, Zheng, Yu, Gao, Huang,
  Lv, et~al.]{qwen3technical2025}
A.~Yang, A.~Li, B.~Yang, B.~Zhang, B.~Hui, B.~Zheng, B.~Yu, C.~Gao, C.~Huang,
  C.~Lv, et~al.
\newblock Qwen3 technical report.
\newblock \emph{arXiv preprint arXiv:2505.09388}, 2025.

\bibitem[Yang et~al.(2026)]{skillopt2026}
Y.~Yang et~al.
\newblock Skillopt: Executive strategy for self-evolving agent skills.
\newblock \emph{arXiv preprint arXiv:2605.23904}, 2026.

\bibitem[Zhang et~al.(2026)Zhang, Zhu, Zhou, Jia, and Wang]{skillevolver2026}
G.~Zhang, E.~Zhu, J.~Zhou, C.~Jia, and H.~Wang.
\newblock Skillevolver: Skill learning as a meta-skill.
\newblock \emph{arXiv preprint arXiv:2605.10500}, 2026.

\bibitem[Zhang et~al.(2025)]{ace2026}
Q.~Zhang et~al.
\newblock Agentic context engineering: Evolving contexts for self-improving
  language models.
\newblock \emph{arXiv preprint arXiv:2510.04618}, 2025.

\bibitem[Zhao et~al.(2023)]{expel2023}
A.~Zhao et~al.
\newblock Expel: Llm agents are experiential learners.
\newblock \emph{arXiv preprint arXiv:2308.10144}, 2023.

\end{thebibliography}

\clearpage
\appendix

\section{Curator Prompt and Output Schemas}
\label{app:prompt}
The curator uses a static benchmark-neutral system prompt and task-specific user prompts that provide the current trace evidence, repository state, and validation summaries. The static prompt used in our experiments is reproduced below.

\begin{Verbatim}[breaklines=true,breakanywhere=true,fontsize=\footnotesize,frame=single]
You are the Curator in Hypothesis-Driven Skill Evolution.

Your responsibility is to improve a frozen Executor by producing narrowly
scoped, evidence-backed task skills. Observe completed Executor traces,
approved skills, the Hypothesis Bank, and prospective paired-validation
results.

Follow this method:

1. Separate execution lapses from recurring, skill-addressable failure modes.
2. Ground every hypothesis in concrete observed trace evidence.
   Require a same-task-type contrast between at least one successful trace and
   one failed trace. A failure alone can diagnose a symptom but cannot establish
   that the proposed policy is better.
3. State an observable applicability condition and explicit risks.
4. Compile concise optional guidance that is safe to ignore when inapplicable.
   The compiled skill must directly operationalize the claimed mechanism.
5. Treat every proposed skill as uncertain until prospective paired validation.
6. In review, reject skills without attributable treatment wins or with
   guardrail regressions, invalid-action regressions, or harmful overreach.
   Separately judge whether the underlying direction is wrong, promising but
   implemented poorly, or merely underpowered. A promising label requires
   paired-trace evidence for a specific correctable defect, not plausibility.
   Any revised skill is a new unvalidated candidate and must be tested again.
7. Explicitly inspect inefficient search, repeated choices, lost multi-stage
   progress, incomplete transformations or reasoning steps, and premature
   assumptions that a goal or answer has been completed.
8. Inspect executor I/O contract failures, including empty outputs, malformed
   outputs, invalid final answers, or actions outside the reported admissible
   action set when the benchmark exposes one. A post-output parser, formatter,
   or resolver can be a valid skill when it is evidence-backed,
   benchmark-neutral, and leaves uncertain cases unchanged.

Do not use benchmark-oracle knowledge, invent hidden environment rules, approve
a skill from a single anecdote, or give broad advice that cannot be validated.
Scope each candidate to the minimum task scope justified by the evidence. A
candidate may cover multiple task types only when the same mechanism is supported
by per-scope trace evidence and the validation plan can test each declared scope.
Treat rejected hypotheses and their paired metrics as negative evidence for
future proposals.
Candidate skill bodies must explain a policy, progress check, or output-contract
repair. Never output a single task answer, environment command, invented action,
or placeholder command such as `look under [object]`.
For reasoning benchmarks, a skill must be an executable reasoning procedure
with a concrete trigger, state/progress check, and decision rule; do not propose
generic reminders such as "think carefully" or "verify the answer" unless the
skill specifies exactly what to verify and how that changes the solution path.
Candidate skills may include pure code payloads for `pre_prompt` or
`post_action` hooks. Code must inspect only provided context fields and must not
encode a task runner, oracle, external calls, or benchmark-specific hidden
knowledge.
For each turn, follow the concrete JSON schema and required fields specified in
the current proposal or review prompt exactly.
\end{Verbatim}

The proposal call supplies compact runtime evidence rather than changing the static system prompt. Its input object has the following shape.

\begin{Verbatim}[breaklines=true,breakanywhere=true,fontsize=\footnotesize,frame=single]
{
  "task_type_outcome_stats": ...,
  "trace_evidence": ...,
  "approved_skills": [...],
  "existing_hypotheses": [...]
}
\end{Verbatim}

The curator must return a JSON array of proposal objects. Each proposal is a falsifiable hypothesis plus one candidate skill package.

\begin{Verbatim}[breaklines=true,breakanywhere=true,fontsize=\footnotesize,frame=single]
{
  "hypothesis_id": string,
  "claim": string,
  "evidence": string,
  "source_trace_ids": [string],
  "attribution": string,
  "applicability_condition": string,
  "expected_improvements": string,
  "guardrails": string,
  "risks": string,
  "falsification_plan": string,
  "promotion_condition": string,
  "rollback_condition": string,
  "confidence": string,
  "candidate_skill": {
    "candidate_id": string,
    "name": string,
    "description": string,
    "body": string,
    "invocation_rule": string,
    "task_types": [string],
    "keywords": [string],
    "payloads": [
      {
        "kind": "text" | "tool" | "code",
        "name": string,
        "content": string,
        "activation": string,
        "entrypoint": string?,
        "safety": string,
        "metadata": object?
      }
    ]?
  }
}
\end{Verbatim}

The review call receives compact hypotheses and paired validation summaries. It is asked to separate the accept/reject decision from the diagnosis of the underlying direction.

\begin{Verbatim}[breaklines=true,breakanywhere=true,fontsize=\footnotesize,frame=single]
{
  "validation_stage": "screening" | "confirmation",
  "hypotheses": [...],
  "paired_validation_results": [
    {
      "metrics_summary": ...,
      "informative_pair_outcomes": ...,
      "behavior_pairs": ...,
      "skill_uptake": ...
    }
  ]
}
\end{Verbatim}

The required review output is:

\begin{Verbatim}[breaklines=true,breakanywhere=true,fontsize=\footnotesize,frame=single]
{
  "hypothesis_id": string,
  "decision": "accept" | "reject",
  "direction_assessment":
    "validated_direction" |
    "promising_needs_revision" |
    "wrong_direction" |
    "insufficient_evidence",
  "failure_attribution":
    "hypothesis" |
    "skill_content" |
    "invocation_scope" |
    "executor_variance" |
    "validation_power" |
    "none",
  "review": string,
  "revision_plan": string,
  "revised_skill": object?
}
\end{Verbatim}

In noisy-feedback experiments, the review prompt states that promotion decisions must use the observed success/failure channel visible to the curator and gate. Hidden oracle metrics are reserved for post-hoc diagnosis and are not included in the curator review input.

\section{Representative Approved Skills}
\label{app:skills}
Below are representative approved skills produced by the curator and copied from the final \method repository. Run-specific metadata is omitted, but the task scope, skill body, invocation rule, payload, and safety condition are preserved.

\begin{Verbatim}[breaklines=true,breakanywhere=true,fontsize=\footnotesize,frame=single]
source: HDSO(27B -> 27B)
candidate_id: skill_heat_direct_action
name: Direct Heating Action Selection
task_types: [pick_heat_then_place]
keywords: [pick_heat_then_place, heating_loop, composite_action, transformation]

description:
  Prioritizes the composite "heat [obj] with [microwave]" action over
  move/open/close sequences to complete heating transformations efficiently.

invocation_rule:
  Trigger during action selection for pick_heat_then_place tasks. Activate when
  holding the target object and current location is the microwave.

body:
  This skill targets an inefficient search and incomplete transformation
  pattern in pick_heat_then_place tasks. The executor frequently moves the
  object into the microwave, then cycles through "open", "close", and
  "examine" actions, failing to trigger the actual heating process. The policy
  introduces a location and inventory check: when the agent is holding the
  target object and arrives at the microwave, suppress "move [obj] to
  microwave", "open microwave", "close microwave", and "examine microwave".
  Instead, directly select "heat [target_obj] with microwave". This composite
  action encapsulates the placement and activation steps, bypassing the
  state-toggle loop. The skill activates only when the target is in inventory
  and the current location matches the microwave. It stops applying once the
  heat action is executed or the task completes. By enforcing this direct
  interaction, the skill aligns with successful trace patterns and eliminates
  redundant navigation and examination steps. Baseline behavior is preserved if
  the agent is not at the microwave or does not hold the target.

payload:
  kind: text
  name: Heating Interaction Guide
  activation:
    Activate when the agent is holding the target object and standing at the
    microwave.
  content:
    1. Check inventory for target object.
    2. Verify current location is "microwave".
    3. If both true, suppress "move", "open", "close", "examine" for microwave.
    4. Queue "heat [obj] with microwave" as highest priority.
    5. Reset state after heating completes or task ends.
  safety:
    Do not apply if the microwave is locked or broken. Do not override if the
    agent is not holding the target.
\end{Verbatim}

\begin{Verbatim}[breaklines=true,breakanywhere=true,fontsize=\footnotesize,frame=single]
source: HDSO(27B -> 27B)
candidate_id: skill_look_at_obj_acquire_first
name: Acquire Target Before Illumination
task_types: [look_at_obj]
keywords: [look_at_obj, desklamp, acquire, loop_prevention, inventory_check]

description:
  Ensures the target object is held before activating a light source to prevent
  examination loops and satisfy task conditions efficiently.

invocation_rule:
  Trigger when task_type is "look_at_obj" and the instruction mentions a light
  source. Activate before selecting actions at any location containing the
  target object or light source.

body:
  When the task requires looking at an object with a light source, check
  inventory first. If the target object is not held, prioritize "take [object]
  from [location]" over "use [light]" or "examine [location]". Once the object
  is confirmed in inventory, proceed to "use [light]". Stop applying this rule
  once the object is held or the light is activated. Preserve baseline action
  selection if the object is already in inventory.

payload:
  kind: text
  name: Acquisition Priority Guide
  activation:
    Activate when the agent is at a location containing the target object or
    light source and has not yet acquired the target.
  content:
    1. Parse instruction for target object and light source.
    2. Query current inventory state.
    3. If target not in inventory and visible in location description, queue
       "take [target] from [location]".
    4. Suppress "use [light]" and "examine [location]" until "take" succeeds.
    5. After successful "take", queue "use [light]".
    6. Reset policy state after light activation or task completion.
  safety:
    Do not force "take" if the object is already in inventory. Do not block
    "use [light]" if the environment indicates the object is already positioned
    correctly.
\end{Verbatim}

\end{document}